\documentclass[conference]{IEEEtran}
\IEEEoverridecommandlockouts
\usepackage{amsmath,amssymb,amsfonts}
\usepackage{algorithmic}
\usepackage{graphicx}
\usepackage{textcomp}
\usepackage{xcolor}
\usepackage{cleveref}
\usepackage{xurl}
\usepackage{bm}
\usepackage{bbm}
\usepackage[countmax]{subfloat}
\usepackage{subcaption}
\usepackage{comment}
\usepackage{microtype}
\usepackage[subtle]{savetrees}

\def\BibTeX{{\rm B\kern-.05em{\sc i\kern-.025em b}\kern-.08em
    T\kern-.1667em\lower.7ex\hbox{E}\kern-.125emX}}
\begin{document}
\bstctlcite{IEEEexample:BSTcontrol}

\title{InTraVisTo: Inside Transformer Visualisation Tool
}

%\author{\IEEEauthorblockN{Anonymous Authors}}
\author{
% \IEEEauthorblockN{Nicolò Brunello}
% \IEEEauthorblockA{\textit{DEIB} \\
% \textit{Politecnico di Milano}\\
% Milan, Italy \\
% nicolo.brunello@polimi.it}
% \and
% \IEEEauthorblockN{Davide Rigamonti}
% \IEEEauthorblockA{\textit{DEIB} \\
% \textit{Politecnico di Milano}\\
% Milan, Italy \\}
% \and
% \IEEEauthorblockN{Andrea Sassella}
% \IEEEauthorblockA{\textit{DEIB} \\
% \textit{Politecnico di Milano}\\
% Milan, Italy \\
% vincenzo.scotti@kit.edu}
% \and
% \IEEEauthorblockN{Vincenzo Scotti}
% \IEEEauthorblockA{\textit{KASTEL} \\
% \textit{Karlsruhe Institute of Technology}\\
% Milan, Italy \\}
% \and
% \IEEEauthorblockN{Mark James Carman}
% \IEEEauthorblockA{\textit{DEIB} \\
% \textit{Politecnico di Milano}\\
% Milan, Italy \\
% mark.carman@polimi.it}
    \IEEEauthorblockN{Nicolò Brunello\textsuperscript{*}, Davide Rigamonti\textsuperscript{*}, Andrea Sassella\textsuperscript{*}, Vincenzo Scotti\textsuperscript{\textdagger}, Mark James Carman\textsuperscript{*}}
    \IEEEauthorblockA{
        \textsuperscript{*}\textit{DEIB}, \textit{Politecnico di Milano}, Milan, Italy \\
        \texttt{\{nicolo.brunello,andrea.sassella,mark.carman\}@polimi.it}, \texttt{davide.rigamonti@mail.polimi.it}}
    \IEEEauthorblockA{
        \textsuperscript{\textdagger}\textit{KASTEL}, \textit{Karlsruhe Institute of Technology}, Karlsruhe, Germany \\
        \texttt{vincenzo.scotti@kit.edu}}
}

\maketitle

\begin{abstract}
The reasoning capabilities of Large Language Models (LLMs) have increased greatly over the last few years, as have their size and complexity. Nonetheless, the use of LLMs in production remains challenging due to their unpredictable nature and discrepancies that can exist between their desired behavior and their actual model output.
In this paper, we introduce a new tool, \textbf{InTraVisTo} (Inside Transformer Visualisation Tool), designed to enable researchers to investigate and trace the computational process that generates each token in a Transformer-based LLM. 
InTraVisTo provides a visualization of both the \emph{internal state} of the Transformer model (by decoding token embeddings at each layer of the model) and the \emph{information flow} between the various components across the different layers of the model (using a Sankey diagram). 
With InTraVisTo, we aim to help researchers and practitioners better understand the computations being performed within the Transformer model and thus to shed some light on internal patterns and reasoning processes employed by LLMs.
\end{abstract}

\begin{IEEEkeywords}
NLP, LLMs, explainability
\end{IEEEkeywords}

\section{Introduction}

The widespread adoption of \emph{Large Language Models} (LLMs) is having a game-changing impact on Natural Language Processing applications in areas from medicine and education, to finance and personal assistants~\cite{DBLP:journals/npjdm/MehandruMASBA24, DBLP:journals/corr/abs-2303-17564}. 
While these large new models come with greatly improved language understanding and text generation skills~\cite{DBLP:journals/corr/abs-2203-15556, DBLP:journals/corr/abs-2303-08774}, the underlying \emph{Transformer Neural Network} they rely on~\cite{DBLP:conf/nips/VaswaniSPUJGKP17}, is still primarily a black-box, in the sense that it is hard to understand exactly \emph{how} the low level data manipulations it performs can lead to the incredible high level reasoning outputs observed. 
This lack of understanding represents a major issue, especially if considered together with the occurrence of  occasional faults in the reasoning process, which lead to imprecise, nonsensical or non-factual outputs -- i.e., ``hallucinations''~\cite{DBLP:journals/csur/JiLFYSXIBMF23}. %-- that we cannot properly explain.
The combination of the lack of interpretability and the presence of errors limits the reliability of LLMs, preventing their deployment in cases where sound reasoning and coherent output are critical. 
Thus, better understanding of the internal reasoning mechanisms and the ability to track down the source of errors are essential to fully exploit LLMs' capabilities. 

\begin{figure}[t]
\begin{center}
    \includegraphics[width=\columnwidth]{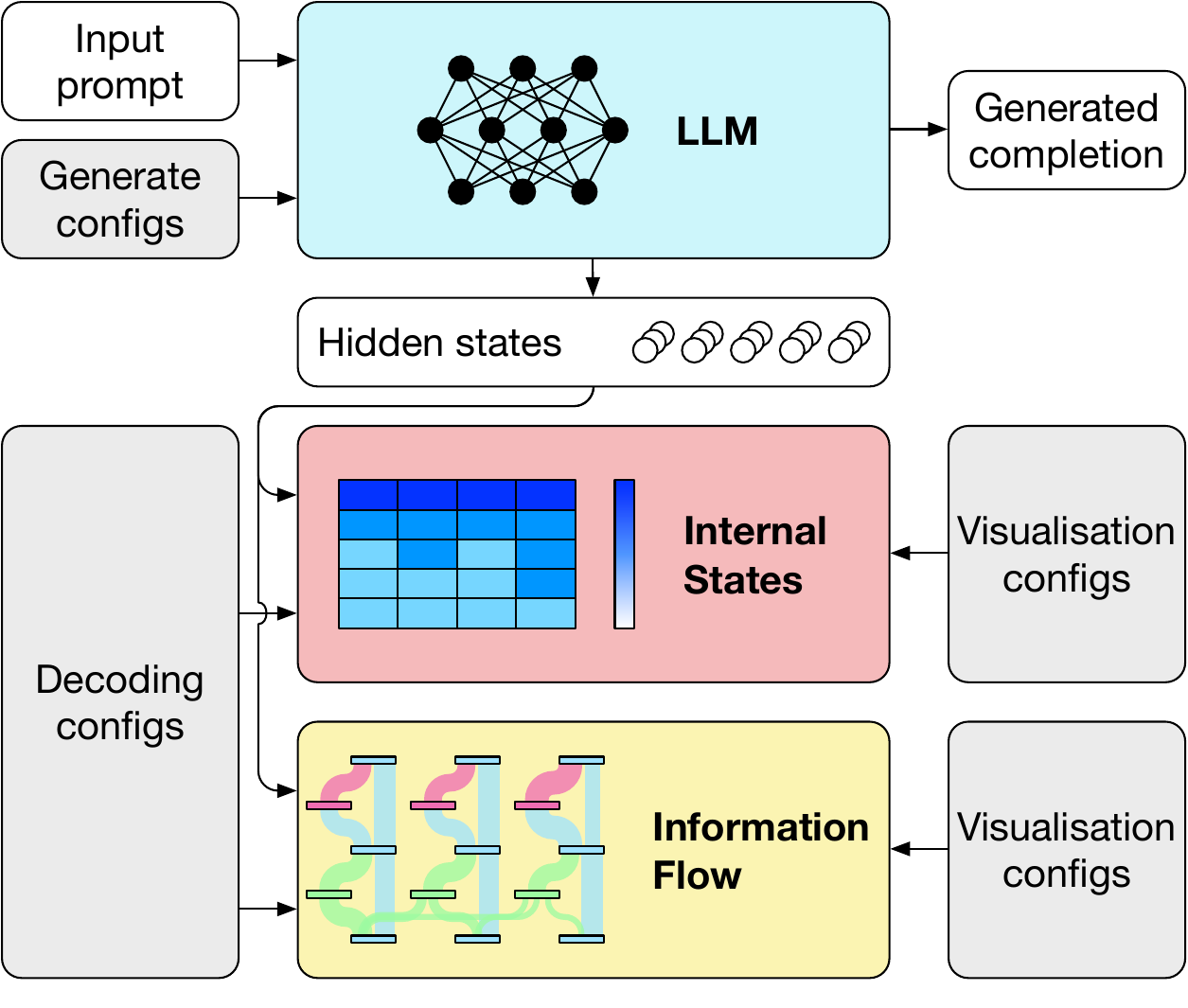}
  \caption{InTraVisTo application overview: given an input prompt and the generate configurations, a LLM generates a completion. We track the hidden (i.e., internal) states produced by the LLM during this process to visualise both these internal states and the information flow using, respectively, a heatmap and a Sankey diagram.}
  \label{fig:general_schema}
\end{center}
\end{figure}

\begin{figure*}[t]  % [!ht]
\begin{center}
    \includegraphics[width=\linewidth]{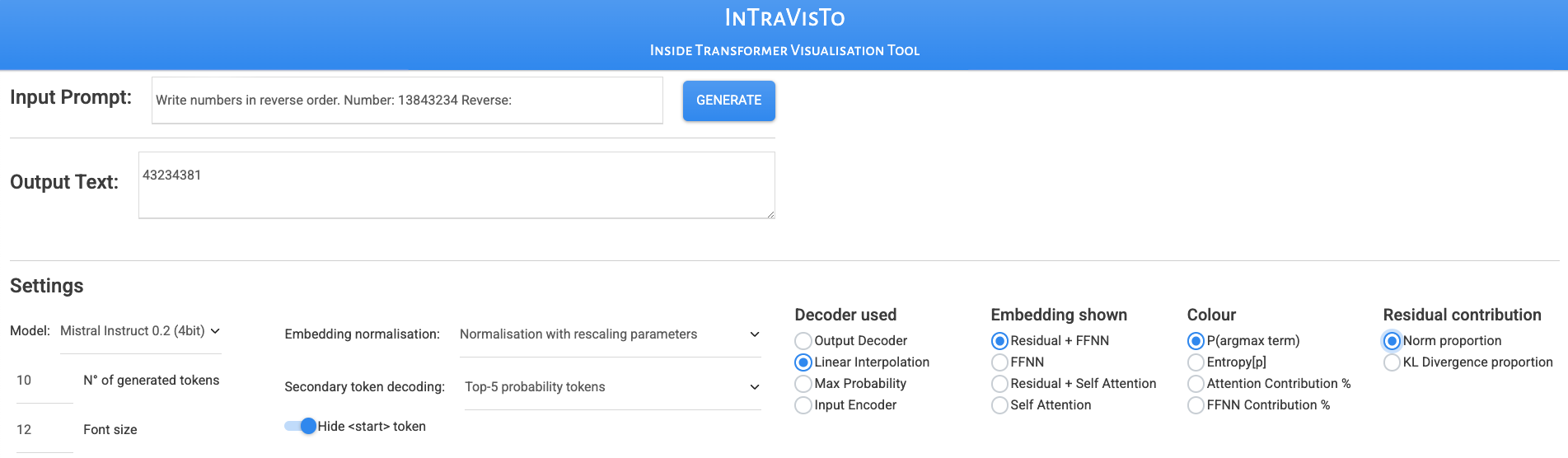} 
    \caption {The main control panel of InTraVisTo. The \emph{Input Prompt} text box on the top left corner allows the user to provide an input to the LLM, the \emph{Generate} button right after starts the generation process on the selected model (the \emph{Model} dropdown allows to change the LLM), and, finally, the \emph{Output Text} box contains the completion to the prompt generated by the LLM. All the remaining controls allow to modify the decoding process and the visualisations.}
    \label{fig:demo_settings}
\end{center}
\end{figure*}

With this goal in mind, we introduce \emph{InTraVisTo}\footnote{``intravisto'' is Italian word meaning ``glimpsed''.} (Inside Transformer Visualisation Tool), which is an open-source visualisation tool\footnote{\url{https://github.com/daviderigamonti/InTraVisTo}}  depicting the internal computations performed within a Transformer.
The tool\footnote{live demo: \url{https://tinyurl.com/2wtudxhf}}, shown schematically in Figure ~\ref{fig:general_schema}, provides visualisations of both the \emph{internal state of the LLM}, using a heat-map of decoded embedding vectors for all layer/token positions, and the \emph{information flow between components of the LLM}, using a Sankey diagram to depict paths through which information accumulates to produce next-token predictions. 

The primary contribution of \emph{InTraVisTo} is its capability to enable researchers and practitioners to interpret and manipulate large language models in real time through a comprehensive set of customizable features, all without requiring any programming expertise.

\section{InTraVisTo}

We now describe the mechanisms and components of the tool. For each main section, we first examine the underlying mathematical implementation, followed by the corresponding application in a use case to offer a practical perspective on the tool and its capabilities. 
Reference \cite{DBLP:conf/nips/LiuAGKZ23} claimed that some model reasoning errors could be tied back to \emph{attention glitches}, minor errors propagated throughout the attention pattern in the model, leading to imperfect state information transferred through the layers. They tested the model on the Flip-Flop language, but a simpler yet effective use case turned out to be \emph{reversing sequence}s.
Thus, the input prompt for our example is: \textit{Write numbers in reverse order. Number: 13843234 Reverse:}. 

\subsection{Decoding Internal States}
\label{sec:decoding}

\begin{figure*}[t]  % 
\begin{center}
    \includegraphics[width=\linewidth]{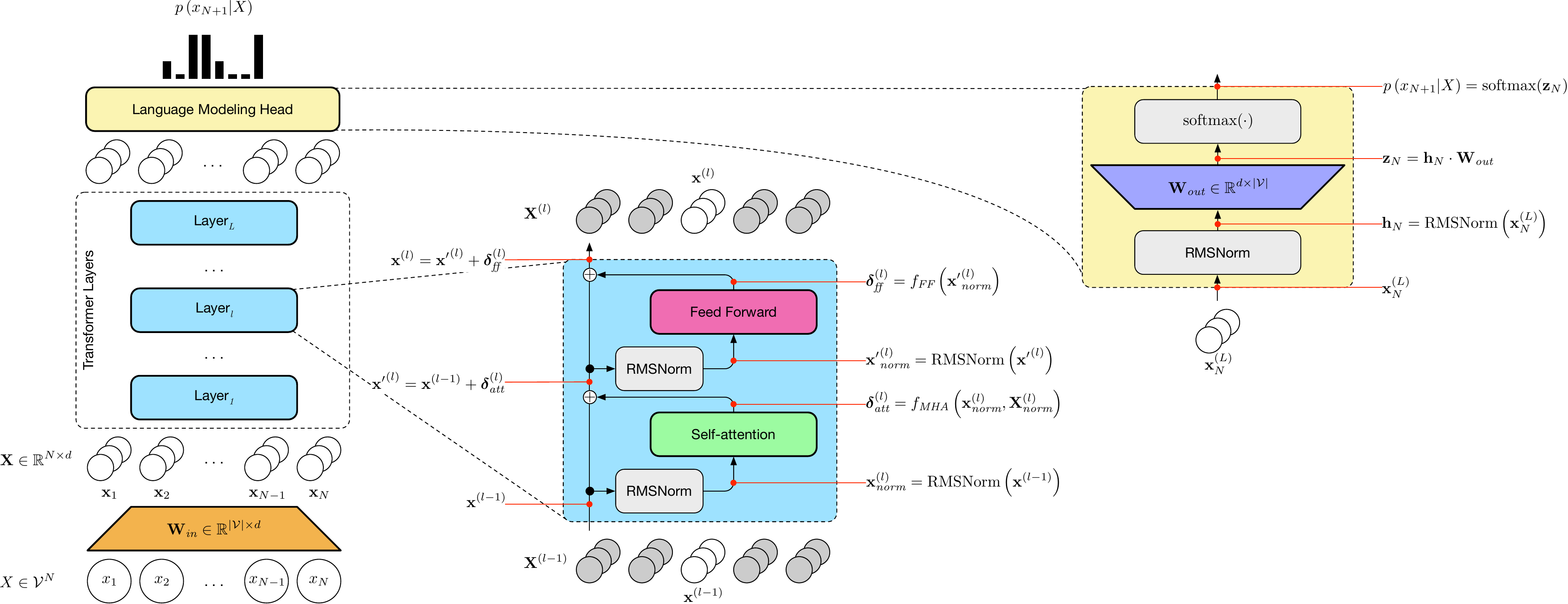}
  \caption{Transformer model architecture. The high-level view of the architecture is represented on the left. In the centre there are the details of a single \emph{transformer layer} and on the right the details of the \emph{language modeling head}.}
  \label{fig:debug_model}
\end{center}
\end{figure*}

Decoding the meaning of hidden state (aka \emph{embedding}) vectors at various depths of a Transformer stack is essential for providing an intuition as to how the model is working.
We focus on causal (i.e., auto-regressive) models, and assume the state-of-the-art LLM architecture shown in \Cref{fig:debug_model}, which involves a \emph{decoration process} whereby each Transformer layer adds the results of its transformations to a residual embedding vector from the layer below. The final embedding vector is then decoded by a language modeling head to produce a probability distribution for the \emph{next token} given the previous tokens.
InTraVisTo provides a human-interpretable representation of this internal decoration pipeline by decoding each hidden state with a specific decoder and displaying the most likely token from the model's vocabulary.

Given a hidden state vector $\mathbf{x} \in \mathbb{R}^d$, which could be the output of a self-attention or feed-forward layer or a residual connection within or outside a transformer block, and a \emph{decoder} matrix $\mathbf{W}_\textit{decoder} \in \mathbb{R}^{d  \times  |\mathcal{V}|}$~\cite{DBLP:journals/corr/abs-2207-09238} consisting of token embeddings, we can \textit{decode} the state by normalising the vector by its RMS value, multiplying by the decoder, and applying the softmax: 
\begin{eqnarray}
P(\textit{token}) = \mathrm{softmax}\left(\frac{\mathbf{x}}{\sqrt{\frac{1}{d}\sum_i x_i^2}} \cdot \mathbf{W}_\textit{decoder}\right) 
\label{eq:softmax}
\end{eqnarray}
to produce a probability distribution over the vocabulary, from which the most likely token is chosen.

Two natural decoders to use are (i) the transpose of the \emph{input embedding matrix} $\mathbf{W}_\textit{in}^\top$ used by the model to convert tokens to vectors on input, and (ii) the \emph{output decoder} $\mathbf{W}_\textit{out}$ used upon output within the language modeling head. 
Some models, like GPT-2~\cite{radford2019language} and Gemma~\cite{DBLP:journals/corr/abs-2403-08295} tie these two parameter matrices together during training ($\mathbf{W}_\textit{in}^\top=\mathbf{W}_\textit{out}$) greatly simplifying the decoding process. 
Popular models like Mistral~\cite{DBLP:journals/corr/abs-2310-06825} and Llama~\cite{DBLP:journals/corr/abs-2307-09288}, however, allow these two matrices to differ ($\mathbf{W}_\textit{in}^\top \neq \mathbf{W}_\textit{out}$), which means that earlier layers tend to be much more interpretable when decoded with the input embedding ($\mathbf{W}_\textit{in}^\top$) while latter layers are more meaningful if the output decoder ($\mathbf{W}_\textit{out}$) is used. 

Previous works have looked to \emph{train specialised decoders}~\cite{DBLP:journals/corr/abs-2303-08112} for this purpose, at the cost of introducing a great deal of additional complexity and potential errors. 
In InTraVisTo we instead make use of a simpler, yet effective, alternative by \emph{interpolating} the input and output decoders based on the depth $l\in[0,L]$ of the model layer we wish to decode:
\begin{eqnarray}
    \mathbf{W}_\textit{linear}^{(l)} =\left(1-\frac{l}{L}\right) \cdot \mathbf{W}_\textit{in}^\top + \frac{l}{L} \cdot \mathbf{W}_\textit{out} 
\label{eq:decoder1}
\end{eqnarray}
A further refinement is made for the case of Llama/Mistral models, which make use of a RMSNorm layer in the language modeling head. Since the layer contains a vector of learnt scaling parameters $\mathbf{s} \in \mathbb{R}^d$, the user is given the option to apply these parameters during the decoding process in Equation~\ref{eq:softmax}. 

Besides these three approaches to decoding, we consider a fourth one for models with separate input and output decoders.
We decode separately with $\mathbf{W}_\textit{in}^\top$ and $\mathbf{W}_\textit{out}$ and we consider the most probable token coming from the two decodings. 
We developed this fourth approach looking at the individual decodings of  $\mathbf{W}_\textit{in}^\top$ and $\mathbf{W}_\textit{out}$ and observing that often (i) the embeddings of the input and output token for a given position coexist in the hidden space throughout the entire stack of contextual representations and (ii) the embedding of the output token is already formed in the first layers with high probability.

\subsection{Visualising Internal States}
\label{sec:heatmap}

\begin{figure*}[t]  % [!ht]
    \centering
    \includegraphics[width=\linewidth]{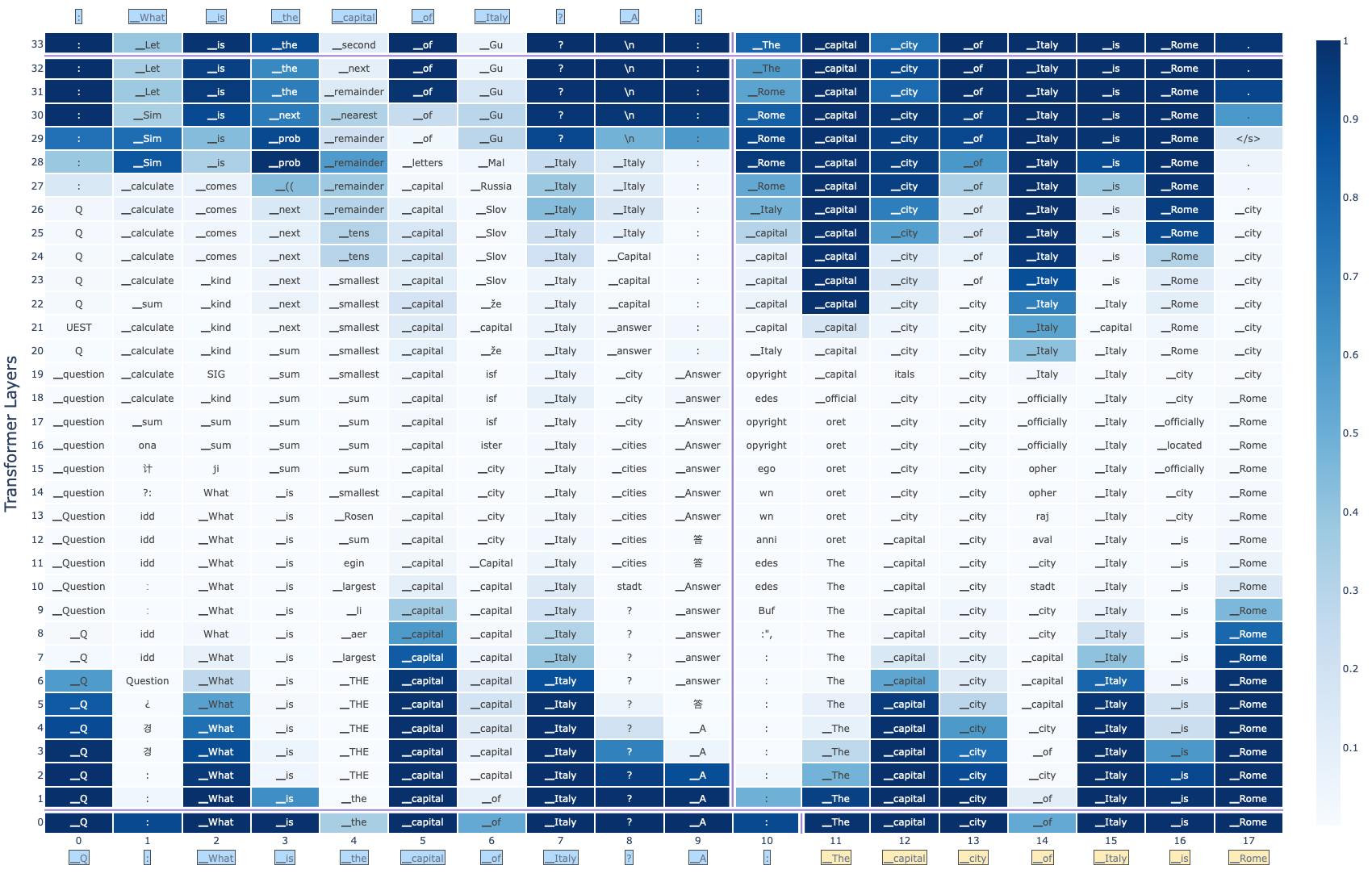}
    \caption{Hidden states heatmap generated by InTraVisTo using the configuration depicted in \Cref{fig:demo_settings}. The input prompt was ``Q: What is the capital of Italy? A:'' with generated completion ``The capital city of Italy is Rome.''}
    \label{fig:heatmap}
\end{figure*}

The first visualisation introduced by InTraVisTo is an annotated heatmap of the decoded internal states (see \Cref{fig:heatmap}). The user can choose a particular decoder (linear interpolation is the default) and views a grid of decoded tokens shaded according to their likelihood for hidden states at each layer and token position of the network. The default hidden state displayed is the output of each transformer block layer (${\mathbf{x}}^{(l)}$), but the user can choose instead to show the change caused by a self-attention block ($\bm{\delta}_\textit{att}^{(l)}$), the change due to the feed-forward layer ($\bm{\delta}_\textit{ff}^{(l)}$), or the residual connection between the self-attention and feed-forward components (${\mathbf{x}'}^{(l)}$). The purpose of the visualisation is to inspect each step in the decoration process, whereby each column corresponds to a token position in the processed sequence and each row corresponds to a specific layer of the Transformer stack. For causal models, the input token is read in at the bottom of each column, and the next token in the sequence is predicted at the top, leading to darker blue cells denoting higher probabilities at both the bottom of each column -- where the input token is being ingested -- and at the top of the column where the model has become more certain of it's prediction for the next token.
A vertical separator indicates the column where the model moves from processing the given input text to generating new output text.

\begin{comment}
\begin{figure}[t]  % [!ht]
    \centering
    % \includegraphics[width=.9\linewidth]{latex/images/intravisto_example_}
    \includegraphics[width=0.3\columnwidth]{images/intravisto_example_wrong_heatmap_crop.pdf}
    \caption{Section of the heatmap grid visualisation in InTraVisTo, focused on the last layers of the output tokens.}
    \label{fig:example_heatmap_wrong}
\end{figure}
\end{comment}

\begin{figure}[t] %[!ht]
\begin{center}
    \centering

    \subfloat[Heatmap slice\label{fig:heatmaperror}]{\includegraphics[width=.88\columnwidth]{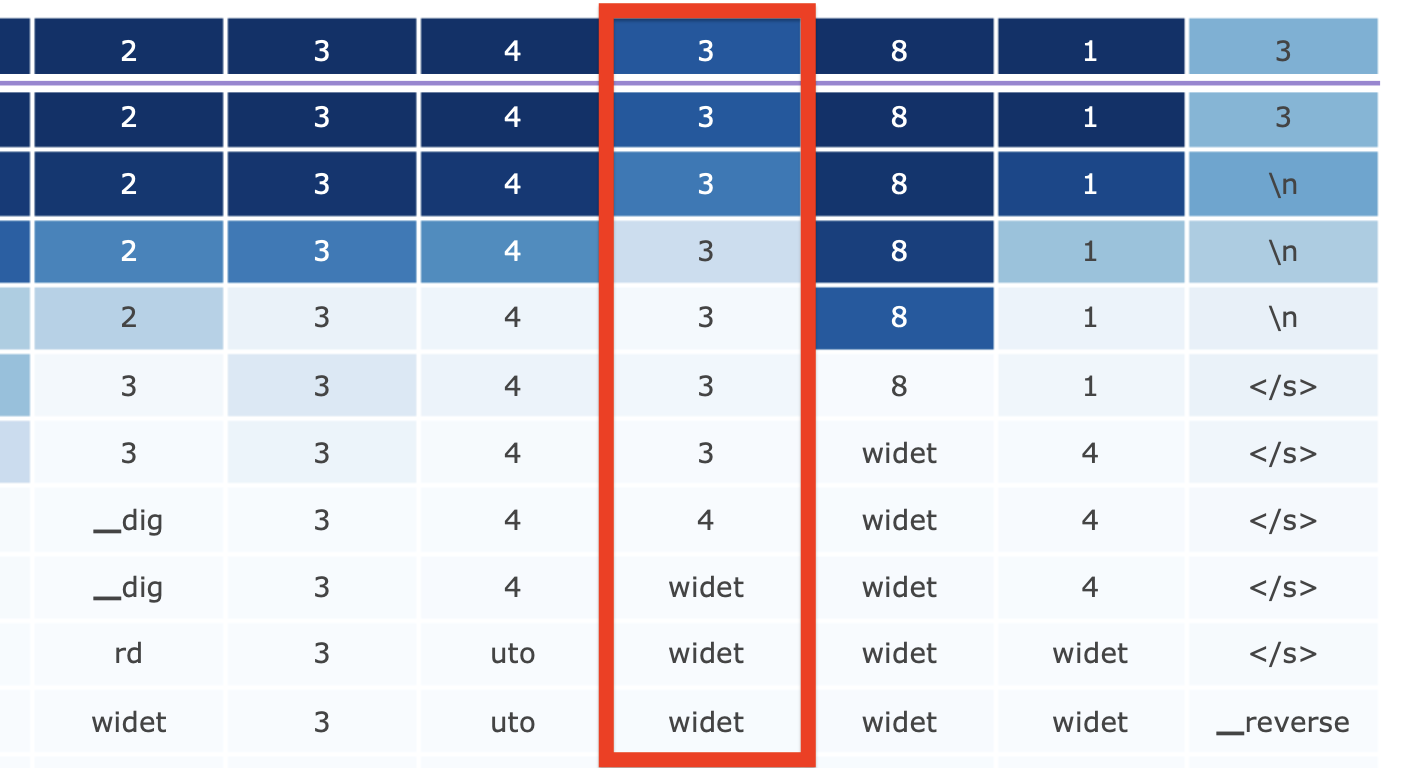}} \\
    \subfloat[$\mathbf{x}^{(l-1)}$\label{fig:heatmapin}]{\includegraphics[width=.16\columnwidth]{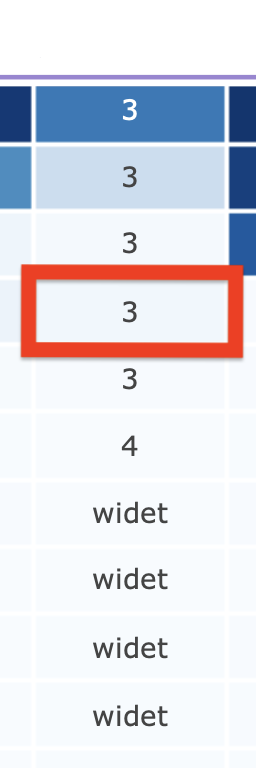}} \hfill
    \subfloat[$\bm{\delta}_\textit{att}^{(l)}$\label{fig:heatmapatt}]{\includegraphics[width=.16\columnwidth]{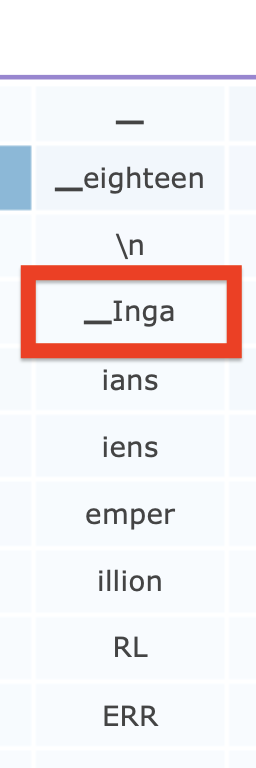}} \hfill
    \subfloat[${\mathbf{x}'}^{(l)}$\label{fig:heatmapintermediate}]{\includegraphics[width=.16\columnwidth]{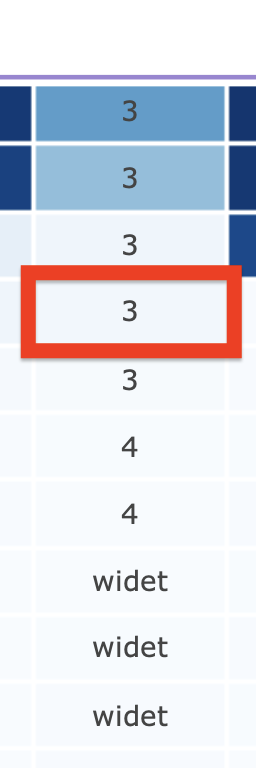}} \hfill
    \subfloat[$\bm{\delta}_\textit{ff}^{(l)}$\label{fig:heatmapff}]{\includegraphics[width=.16\columnwidth]{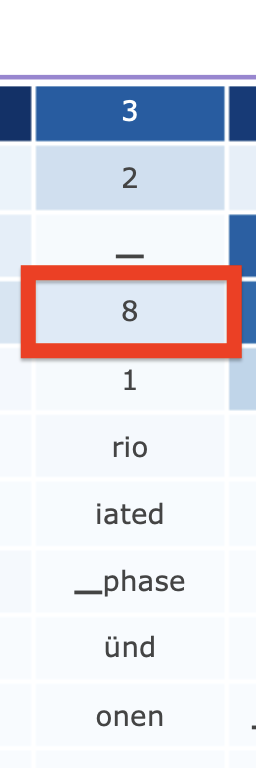}} \hfill
    \subfloat[$\mathbf{x}^{(l)}$\label{fig:heatmapout}]{\includegraphics[width=.16\columnwidth]{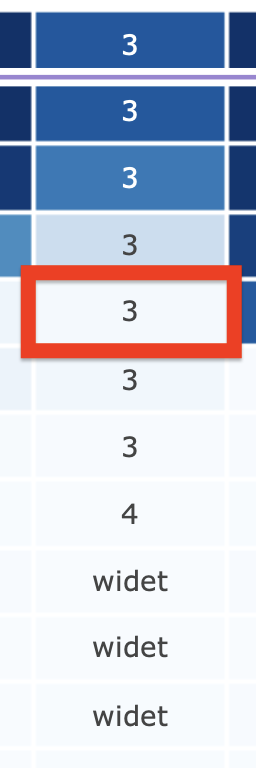}}
\caption{Slice of a heatmap from erroneous generation with details of the error decoding different hidden vectors. The input prompt was
``Write numbers in reverse order. Number:
13843234 Reverse:'' with erroneous generated completion ``43234381''. Despite the FFNN tries to push the correct digit ``8'' on the residual stack the wrong output digit ``3'' is not overwritten in the intermediate representation and stays up to the output.}
\label{fig:heatmapdetail}
\end{center}
\end{figure}

%In \Cref{fig:heatmap}, the LLM was prompted with the question \emph{``What is the capital of Italy?''}. The visualisation of decoded internal states helps us to follow the answer generation process, which appears to resemble some form of reasoning process. The model initially produces an embedding of the token \texttt{\_Rome} in layer 27 of position 10 (immediately following the input text), but then proceeds to generate a more complete response \emph{``The capital city of ...''}, with the token \texttt{\_Rome} postponed to position 16.
%We report an additional noteworthy example of use case in \Cref{sec:appendix}.

%As noted above, the grid forms a heatmap, where the colour of each cell denotes the \emph{probability} of the displayed token, indicating how \emph{confident} the model is at that point. 
The grid forms a heatmap, where the colour of each cell denotes the \emph{probability} of the displayed token, but the user can choose to base the colour on the \emph{entropy} of the distribution, indicating how \emph{undecided} the model is. Moreover, in order to visualise the \emph{importance} of the multi-head \emph{self-attention} or the \emph{feed-forward} neural network at each layer, the colour can be based on the \emph{relative contribution} of these components, calculated using the norm of the respective vectors: %(Equations~\ref{eq:mhacontribution} \&~\ref{eq:ffnncontribution}).
\begin{eqnarray}
    {\%}^{(l)}_{\textit{attention}} = \frac{\|\bm{\delta}_\textit{att}^{(l)}\|_2}{\|\bm{\delta}_\textit{att}^{(l)}\|_2 + \|\mathbf{x}^{(l-1)}\|_2} 
\label{eq:mhacontribution} \\
    {\%}^{(l)}_{\textit{feed-forward}} = \frac{\|\bm{\delta}_\textit{ff}^{(l)}\|_2}{\|\bm{\delta}_\textit{ff}^{(l)}\|_2 + \|{\mathbf{x}'}^{(l)}\|_2}
\label{eq:ffnncontribution} 
\end{eqnarray}

\begin{comment}
Referring to our illustrative example, as depicted in \Cref{fig:example_heatmap_wrong}, the output of the model is not correct as it is \emph{43234\textbf{38}1} instead of \emph{43234\textbf{83}1}. Observing \Cref{fig:heatmap}, it is immediately clear that these models (in this case \texttt{mistralai/Mistral-7B-Instruct-v0.2}) present a clear step in the magnitude of probabilities when passing from the intermediate layers to the last layers. Through manual trials, we saw that this step tends to happen earlier in the layers as the next token becomes more obvious. In this case, as depicted in \Cref{fig:example_heatmap_wrong}, the layer at which this step happens for the wrongly generated digit appears to be one layer late, with respect to the other correct digits. This behaviour is a symptom that the model is ``not so sure'' about this generation.
This statement seems corroborated by the magnitude of the probabilities for the wrong token, that even in the last layers, seems to be lower (lighter background color) compared with the correct token probabilities.  
Knowing that in \Cref{fig:example_heatmap_wrong} the correct final token in the red highlighted box should be \texttt{8} instead of \texttt{3}, let's now check if the model has ever introduced an \texttt{8} during its reasonings.
\end{comment}
Referring to our example of the number digit inversion, as depicted in \Cref{fig:heatmapdetail}, the output of the model is not correct as it is \emph{43234\textbf{38}1} instead of \emph{43234\textbf{83}1}. 
Observing \Cref{fig:heatmaperror}, it is immediately clear that these models (in this case \texttt{mistralai/Mistral-7B-Instruct-v0.2}) present a clear step in the magnitude of probabilities when passing from the intermediate layers to the last layers. 
Through manual trials, we saw that this step tends to happen earlier in the layers as the next token becomes more obvious. 
In this case, as depicted in \Cref{fig:heatmaperror}, the layer at which this step happens for the wrongly generated digit appears to be one layer late, with respect to the other correct digits. This behaviour is a symptom that the model is \emph{not confident} about this generation.
This statement seems corroborated by the magnitude of the probabilities for the wrong token, that even in the last layers, seems to be lower (lighter background color) compared with the correct token probabilities.  
Knowing that in \Cref{fig:heatmaperror} the correct final token, which appears on top of \Cref{fig:heatmapout}, should be \texttt{8} instead of \texttt{3}, we can check whether the model has ever introduced an \texttt{8} during its ``reasoning''. 

\begin{comment}
\begin{figure}[t]  % [!ht]
    \centering
    % \includegraphics[width=.9\linewidth]{latex/images/intravisto_example_}
    \includegraphics[width=0.5\columnwidth]{images/intravisto_example_multiple_hetmap.pdf}
    \caption{Same section of the heatmap highlighted in \Cref{fig:example_heatmap_wrong}, but considering different components for decoding: attention (green) residual plus attention (yellow), feed forward (blue).}
    \label{fig:example_heatmap_multiple}
\end{figure}
\end{comment}

% From \Cref{fig:example_heatmap_multiple} we can see that despite the tokens decoded from the attention and residual plus attention embeddings are not so informative, in the ones coming from the FF, there is an \texttt{8} in the fourth-last layer (see \Cref{}). Now we can conjecture that the model result was probably wrong due to an incorrect or imprecise internal representation around the $29^{th}$ layer, that failed to propagate the correct information above. In the next section we will delve more into the example, while showing the functionalities of InTraVisTo.

From \Cref{fig:heatmapatt,fig:heatmapintermediate,fig:heatmapff} we can see that, despite the tokens decoded from the attention and residual plus attention embeddings (see respectively \Cref{fig:heatmapatt,fig:heatmapintermediate}) are not so informative, in the embeddings coming from the FFNN, there is an \texttt{8} in the fourth-last layer (see \Cref{fig:heatmapff}). Now we can hypotesise that the model result was probably wrong due to an incorrect or imprecise internal representation around the $29^{th}$ layer, that failed to propagate the correct information above. % In the next section we will delve more into the example, while showing the functionalities of InTraVisTo.

\subsection{Visualising Information Flows}
\label{sec:sankey}

\begin{figure*}[t]  % [!ht]
    \centering
    \includegraphics[width=\linewidth]{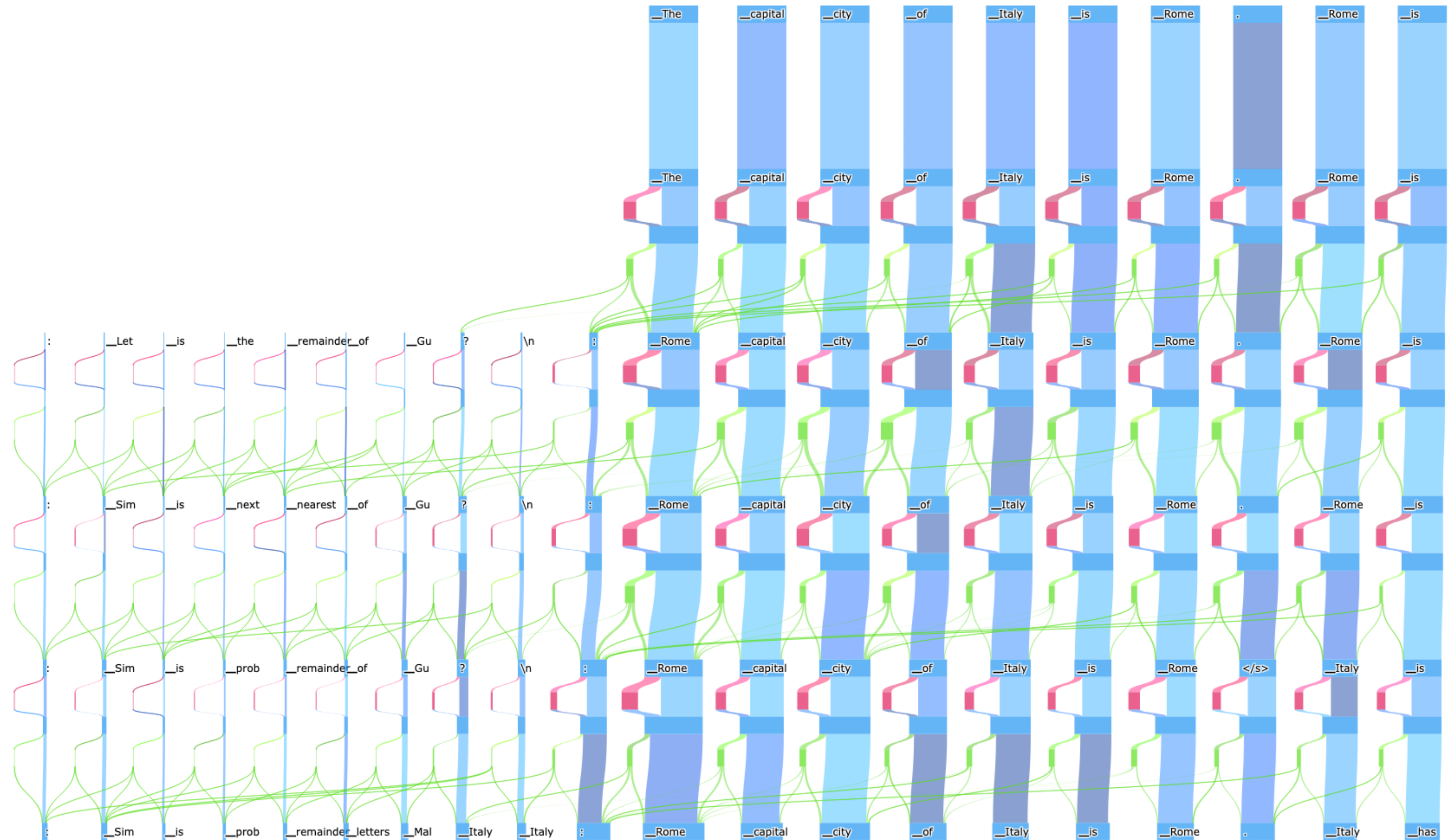}
    \caption{Top 5 layers of a Sankey diagram generated by InTraVisTo using the configuration depicted in \Cref{fig:demo_settings}. The input prompt was ``Q: What is the capital of Italy? A:'' with generated completion ``The capital city of Italy is Rome.'' (blue is the residual connection flow, green is the attention layers flow and pink is the FF layers flow).}
    \label{fig:sankey}
\end{figure*}

\begin{figure}[t]  % [!ht]
    \centering
    \includegraphics[width=\columnwidth]{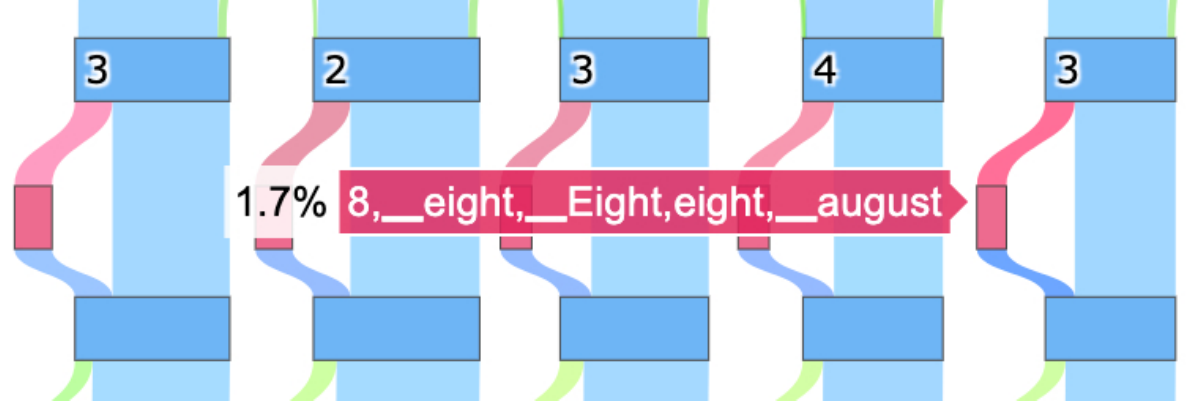}
    \caption{Flow percentages visualised by hovering the mouse above the desired block in the Sankey diagram representation. The input prompt was
    ``Write numbers in reverse order. Number:
    13843234 Reverse:'' with erroneous generated completion ``43234381''. Despite the FFNN tries to push the correct digit ``8'' on the residual stack the wrong output digit ``3'' is not overwritten in the intermediate representation and stays up to the output.}
    \label{fig:example_flow_perc}
\end{figure} 
The second visualisation introduced in InTraVisTo is a Sankey diagram that aims to depict the information flow through the Transformer network (\Cref{fig:sankey}). Edges in the diagram indicate the amount of influence that the nodes have on each other and show how the information accumulates from the bottom of the diagram to the top in order to generate the final prediction. The flow snakes its way through self-attention nodes (in green), which combine information from attended tokens in the level below, feed-forward networks nodes (in pink), which introduce information based on detected patterns in the state vector, and aggregation nodes (in blue), where updates from the other two types of nodes are added to the residual vector. A complete Sankey diagram containing 32 layers of Transformer blocks would be very tall, so only the last 5 layers of the network are shown in \Cref{fig:sankey}.

In order to calculate the information flow, an attribution algorithm works backwards from the top layers of the network, recursively apportioning the incident flow from the components below based on their relative contributions to the internal state vector above (using the Equations~\ref{eq:mhacontribution} \&~\ref{eq:ffnncontribution}). The calculations for the feed-forward, intermediate aggregation and attention nodes are then simply:
\begin{eqnarray}
    \textit{flow}_{\textit{ffnn}}^{(l,j)} &=& {\%}_{\textit{ffnn}}^{(l,j)}*\textit{flow}_{x}^{(l,j)}\\
    \textit{flow}_{x'}^{(l,j)} %&=& \textit{flow}_{\textit{ffnn}}^{(l,j)}  + ( 1 - {\%}_{\textit{ffnn}}^{(l,j)})* \textit{flow}_{\textit{x}}^{(l,j)} \nonumber \\
                               &=& \textit{flow}_{x}^{(l,j)}\\
    \textit{flow}_{\textit{att}}^{(l,j)} &=& {\%}_{\textit{att}}^{(l,j)}*\textit{flow}_{x'}^{(l,j)}
\end{eqnarray}
Computing the outgoing flow from the aggregation node at the next layer down ($l\!-\!1$), requires combining all outgoing contributions to subsequent self-attention nodes as well as the residual flow:

\begin{equation}
    % \textit{flow}_{x}^{(l-1,j)} = \sum_{i\in[j,k]}\overline{\textit{attend}}^{(l,i)}[j]*\textit{flow}_{\textit{att}}^{(l,i)} \nonumber + ( 1 - {\%}_{\textit{att}}^{(l,j)})* \textit{flow}_{\textit{x'}}^{(l,j)} 
    \textit{flow}_{x}^{(l-1,j)} = \sum_{i\in[j,k]}\mu_\textit{att}^{(l,i)}[j]*\textit{flow}_{\textit{att}}^{(l,i)} + ( 1 - {\%}_{\textit{att}}^{(l,j)})* \textit{flow}_{\textit{x'}}^{(l,j)} 
\end{equation}

Where $\mu_\textit{att}^{(l,i)}[j]$ denotes the average attention placed on position $j$ by the attention heads at level $l$, position $i$.

The aim of the diagram is to visualise the relative contributions of the self-attention and feed-forward networks to the residual connection between layers, and thus allow the user to track down the sources of the major contributions to the prediction of the output at the top of the diagram.

As an alternative to the attribution of importance based on the relative size of the vectors (given by Equations~\ref{eq:mhacontribution} \& ~\ref{eq:ffnncontribution}), we consider also enabling the scaling of the flows according to the similarity between the resulting term distributions, whereby similar distributions are apportioned more weight since they are more likely to have been the ``source of the information'' flowing upwards in the network. The \emph{KL-Divergence} is used to compute the similarity between the distributions on both branches and then normalised to be used as a node contribution weight. 

Oftentimes, the complexity of the attention patterns can make the Sankey diagram difficult to read. To cope with this problem, we enable the filtering of the edges to show only the top $k$ most significant contributions to the attention, discarding the remaining ones, making it easier to visually track the incoming information in the final embedding.
Finally, we decorated the nodes showing the $k$ most probable tokens obtained decoding the current embedding (to observe the neighborhood of the embedding) and the $k$ most probable tokens obtained difference between the output embedding and the input embedding to that node (to observe the information being added to the flow by that node).
As an alternatives, we also considered an \emph{iterative decoding strategy} consisting in iteratively removing the components of the most probable token from the hidden state, generating a new decoded embedding for each iteration until the norm of the state falls under a certain threshold or a fixed number of iterations is reached.
We can now apply information flow analysis to our running example. By inspecting the information flow representation in \Cref{fig:example_flow_perc}, we can see that the flow percentages of the FF block at the $29^{th}$ layer highlighted before is 1.7\%, while the other contribution at the same layer are at least 1.8\%. This indicates that that particular block is slightly below the average in terms of contributions to the overall information flow. 

From these considerations, it seems that at a certain point, the model probably makes the correct computation, but actually forgets the final target of the task (i.e. reversing the sequence). We can check this conjecture in the next section by injecting an embedding and probe the behaviour of the model. 

\subsection{Injection of embeddings}
\label{sec:injection}
We introduce \emph{embedding injection} functionality to further understand the internal mechanisms of the LLM and to enable some form of interaction and probing the model. It allows the user to \emph{substitute} the hidden state vector at any position and depth with the embedding of a token chosen from the vocabulary. This is, to the best of our knowledge, a feature that was not explored in the current state of the art tools. 

The main approach we consider to implement injection is by removing a component $\mathbf{e}_\textit{old}$ of an hidden state $\mathbf{h}$ (whether it is $\mathbf{x}^{(l)}$, ${\mathbf{x}'}^{(l)}$, $\bm{\delta}_\textit{att}^{(l)}$ or $\bm{\delta}_\textit{ff}^{(l)}$) and replacing it with another vector $\mathbf{e}_\textit{new}$, with $\mathbf{e}_\textit{old}$ and $\mathbf{e}_\textit{new}$ being normalised embeddings from $\mathbf{W}_\textit{decoder}$.
We compute the updated hidden vector $\mathbf{h}_\textit{inject}$ --taking care of scaling the components to remove and to add by the correlation between $\mathbf{e}_\textit{old}$ and $\mathbf{h}$-- as in the following formula:

\begin{equation}
    \mathbf{h}_\textit{inject} = \mathbf{h} + \mathbf{h} \cdot \mathbf{e}_\textit{old} \cdot (\mathbf{e}_\textit{new} - \mathbf{e}_\textit{old})
\end{equation}

\noindent
Besides this implementation of the injection, we consider also alternative approaches, by allowing to optionally to do an complete replacement of the hidden vector (rather than only the main component) or by disabling the scaling of the components.

\begin{figure}[t]  % [!ht]
    \centering
    \includegraphics[width=.88\columnwidth]{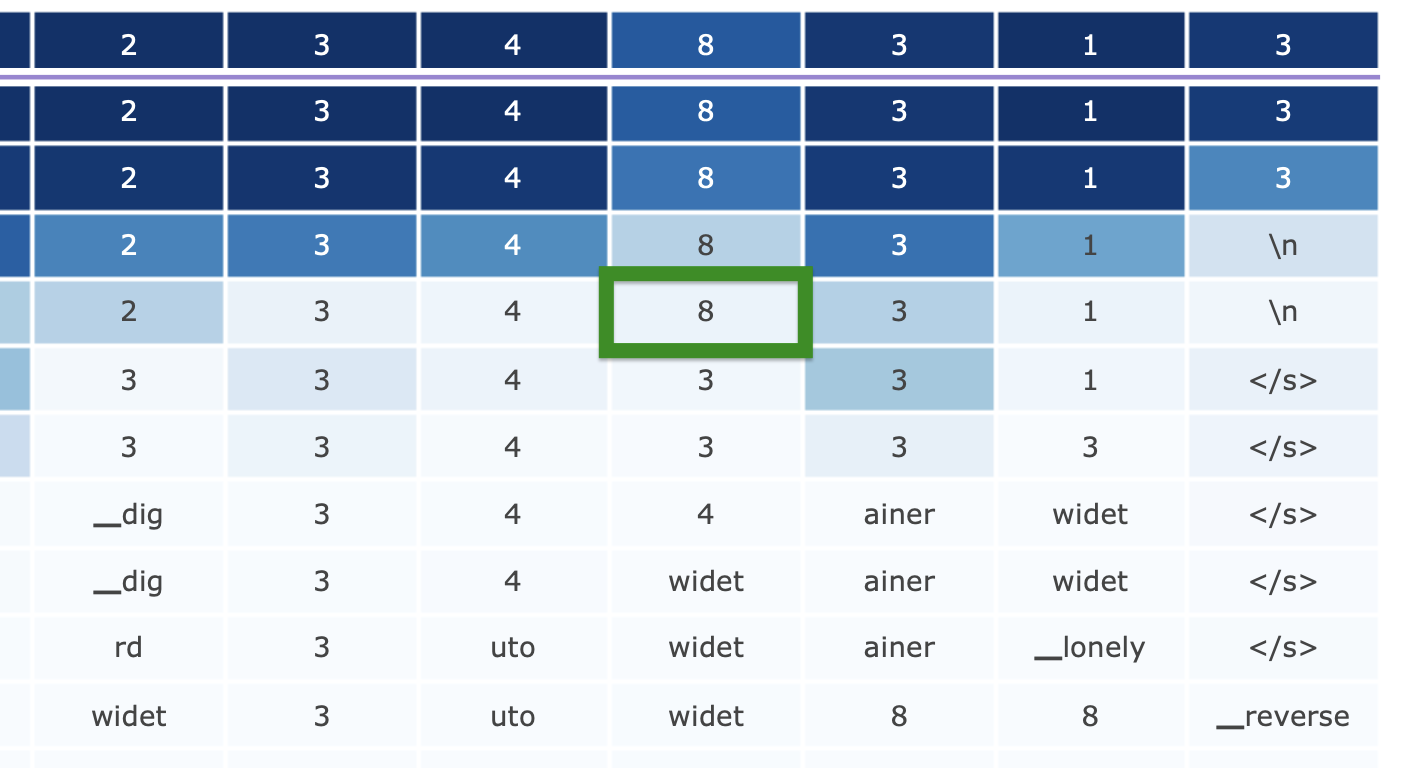}
    \caption{Slice of the heatmap around injected embedding of ``8'' (highlighted in green) to replace that of ``3'' (see \Cref{fig:heatmaperror} for original) The input prompt was
``Write numbers in reverse order. Number:
13843234 Reverse:'' with correct generated completion ``43234831''. Before injecting the ``8'' the erroneous completion was ``43234381''.}
    \label{fig:example_heatmap_inj}
\end{figure}  

% The injection procedure starts when the user clicks on a cell in the heatmap, writes in the pop-up block the token to inject and presses again the generation button. %The generation process runs as before and redraws all the components, but this time while it is generating the embedding injecting is substituted to the real one at the specified location. 
Coming to our number inversion example, we see the effect of injecting an \texttt{8} before the erroneous prediction of the \texttt{3} generating, thus, the correct output \emph{432348313} (see \Cref{fig:example_heatmap_inj}). 
At first, it seems that the problem concerning the two swapped digits is solved, due to the fact that now the \texttt{8} and the \texttt{3} are in the correct positions, but this time the model added a digit at the end of the generation, instead of the new line character as before. By inspecting the new heatmap in \Cref{fig:example_heatmap_inj} the user can notice that the probability of token \texttt{8} is notably increased. Moreover, concerning the last wrong token, it is also clear that in the lower layer the model converged to the new line token (probably due to the end of the length of the sequence), but as the layer level reaches and is above the level of the injection, the output changed to a sub-optimal token with respect to the final task. This could be expected since we are injecting an embedding that was not crafted from the network at that particular layer, so there could be minor modifications that, through causal attention, are influencing the next tokens negatively.
\section{Related works}
\label{sec:background}

Hereafter, we explore related works on LLMs interpretability and explainability, focusing on \emph{decoding within the vocabulary space}~\cite{DBLP:journals/corr/abs-2405-00208}.

\subsection{Hidden State Intepretation}
\label{sec:hiddeninterpret}

\emph{Logit Lens}~\cite{nostalgebraist2020logitlens} represents the first major step towards the decoding and interpretation of a Transformers hidden states. Logit Lens showed examples of decoding the internal sates of GPT-2 using the embedding matrix (which is shared between the input and the output). The results, as in the case of InTraVisTo, is a probability distribution over the vocabulary. From this distribution, it was possible to extract the most probable token(s) or look at the confidence of the predictions.

%This leads to the discovery of some interesting properties, such as the fact that the model almost immediately (i.e. in the first layers) starts generating the output token inside the hidden state representation, and the possible emergence inside hidden states of rare tokens that were encountered in the previous portion of the computed sentence.

%Other interesting developments following the concept of \textit{logit lens} exist, such as the \textit{tuned lens}~\cite{DBLP:journals/corr/abs-2303-08112}, the \textit{attention lens}~\cite{DBLP:journals/corr/abs-2310-16270}  and the \textit{future lens}~\cite{DBLP:conf/conll/PalSYWB23}.
%These derived techniques are brought together by the usage of probes~\cite{DBLP:conf/iclr/AlainB17} to enhance the interpretation of hidden representations, opening up the possibility of extracting targeted information about specific elements of the computation.

Logits Lens was the basis for other approaches based on \emph{probing}~\cite{DBLP:journals/corr/abs-2310-16270,DBLP:journals/corr/abs-2303-08112,DBLP:conf/conll/PalSYWB23}, which refers to the use of linear projection crafted to observe the internal states of a neural network~\cite{DBLP:conf/iclr/AlainB17}. Among these probing approaches, \emph{Tuned Lens}~\cite{DBLP:journals/corr/abs-2303-08112} is one of the most noteworthy. 
% In fact, Tuned Lens introduced a translation transformation that correct for rotation, scale and bias of hidden states with linear transformation to help the decoding.
In fact, Tuned Lens introduced a linear translating transformation of hidden states to achieve the decoding.
The parameters of the linear transformation for a layer $l$ are obtained optimising a distillation loss measuring the KL-Divergence of the final probability distribution from the decoded distribution of $l$. Moreover,~\cite{DBLP:journals/corr/abs-2303-08112} argued how properly debiasing and normalising the hidden states influenced the meaningfulness of the decoding.
We deliberately chose not to incorporate a pre-trained translator for two key reasons. First, doing so would limit the applicability of \emph{InTraVisTo} to only those models equipped with a trained translator. Second, our primal and novel objective was to evaluate the model's self-interpretability capabilities and determine the extent to which a single model can provide insights into its own internal workings.

\subsection{Information Flow Visualisation}
\label{sec:flowvisual}

Analysing the information flow inside a transformer is a challenging a non-trivial task. In this context, the attention weights play a significant role~\cite{DBLP:conf/acl/Vig19,DBLP:conf/acl/WangTC21,DBLP:journals/corr/abs-2404-07004}. 
Early approaches, like \emph{BERTViz}~\cite{DBLP:conf/acl/Vig19}, focused on visualising the attention weights within layers individually. These static visualisations proved useful to notice that the Transformer heads were specialised in some tasks, like co-reference resolution. More sophisticated solution, like \emph{Dodrio}~\cite{DBLP:conf/acl/WangTC21}, extend this visualisation trying to provide better interpretations of the role of different heads.
To the best of our knowledge, existing tools do not provide the same level of insight that we aim to achieve with \emph{InTraVisTo}. Specifically, these tools fail to reveal the \emph{relative contribution} of each component. Instead, they focus on interpreting individual attention heads and visualizing the flow of information based on attention patterns within the network. Consequently, their visualizations do not allow users to precisely identify the specific component (in terms of layer and position) responsible for generating a particular output.

The \textit{LM Transparency Tool} (LM-TT)~\cite{DBLP:journals/corr/abs-2404-07004}, on the other hand, focuses on the big picture.
LM-TT builds on top the \emph{Circuit Transformers} interpretation to visualise the information flow. In the Circuit Transfomer,~\cite{elhage2021mathematical} ran an extensive ablation how the self-attention and the feed forwards throughout transformer layers work to compose information exploiting the residual connection and communicate using some dimensions of the embedding vectors.
LM-TT tool focuses on the visualisation of the most relevant attention paths leading to the production of an embedding in the internal states of the transformer. In this regard, mechanistic interpretability has shown remarkable progress towards identifying circuits and interpreting their functionalities \cite{DBLP:conf/iclr/WangVCSS23}, nevertheless we resort to a more minimal flow interpretation due to the high computational cost of activation and path patching experiments.

\subsection{Information Injection}
\label{sec:infoinj}

Information injection aims at understanding the behaviour of a LLM by altering the internal states. %  during inference. 
Conceptually, this approach in explaining a model behaviour can be seen as the white-box equivalent of black-box explanations obtained injecting random noise in the model input, like those obtained with \emph{LIME}~\cite{DBLP:conf/kdd/Ribeiro0G16}. 
%\cite{DBLP:conf/nips/GeigerLIP21} first explored these approaches is a structured way with \emph{causal interventions}, a framework to characterise the role a LLM internal representations.
\cite{DBLP:conf/nips/GeigerLIP21} first explored these approaches is a structured way introducing the \emph{causality framework} to transformer architectures, with the purpose of characterising the roles of the internal components of a model.
This allowed the use of \emph{activation patching}, which consists in changing the values of the computations performed by the models' components.

%The concept of performing causal interventions has already been explored extensively in the current literature.
Frameworks like \textit{Patchscopes}~\cite{DBLP:journals/corr/abs-2401-06102} incorporate the possibility of performing causal interventions on language models. %  in the context of vocabulary space decoding. 
%\textbf{TODO explain activation patching?}
Other frameworks applying similar activation patching techniques include \textit{Garçon}~\cite{elhage2021garcon} and \textit{TransformerLens}~\cite{nanda2022transformerlens}.
In particular, \textit{TransformerLens} serves as as backend in many interpretability visual applications including Tuned Lens and LM-TT, which have been treated previously.
% , providing an insightful reference point for our analysis.
Other causal intervention techniques~\cite{DBLP:conf/nips/0001LV23, DBLP:conf/nips/ConmyMLHG23, DBLP:conf/iclr/WangVCSS23} focus on the acquisition of insights by programmatically computing the difference between ``clean'' and ``patched'' runs. 
% \textbf{TODO explain main difference from previous frameworks}
% In our work we focus on providing an activation patching interface that is completely functional to the main visualizations, and not centered around a formal causal analysis approach.
In our work we focus on a patching interface that is completely functional to the main visualizations, and not centered around a formal causal analysis approach. To the best of our knowledge, \emph{InTraVisTo} is the only tool that enables the modification of embeddings within the hidden states of a network through a graphical interface. This unique capability allowed us to explore various injection modalities, which are made accessible to users through selectable options in the interface, such as the choice of normalization techniques or whether to apply normalization at all.
%Its main purpose is to aid the user in the interactive exploration of model behavior.
% and a  one through the use of a $\textrm{diff}(\cdot,\cdot)$ function.
% In our case, we choose to implement interventions in a purely user-oriented way by performing activation patching with a properly encoded, user-provided injection prompt.
% Additionally, due to the nature of our implementation, we do not put restrictions on the computation of the difference between `patched' and `clean' runs, leaving freedom of interpretation to the user.

% Most tools that enable activation patching are presented as libraries, providing support for conducting experiments through the approach of causal interventions.

\subsection{Discussion of Related Works}
\label{sec:discussion}

InTraVisTo combines into a single package the functionalities of (i) interpreting the hidden state (without the need to train any decoder), (ii) observing the information flow, and (iii) allowing interactive injection of modified information. 
% It does so while providing a great deal of additional visualisation options, all in order to allow for a more fine-grained analysis and interactive debugging of an LLM
Moreover, our tool offers a wide range of options to customize the visualisation allowing for a more fine-grained analysis and interactive debugging of the LLM being examined.

To the best of our knowledge, no other tool covers at once all these functionalities.
All the tools we presented in this section mainly focus on one aspect, like hidden states visualisation or information flow visualisation, losing some information that may be crucial in understanding the internal process of the LLM.

\section{Conclusion}

We presented InTraVisTo, a tool to visualise and manipulate the internal states and the information flow of Transformer Neural Network, the core of LLMs. 
We developed \emph{InTraVisTo} to assist researchers and NLP practitioners in identifying potential flaws within the architecture of LLMs and gaining a deeper understanding of how these models generate subsequent tokens. 
Ultimately, we aim for our tool to enhance the reliability of LLMs by enabling individuals without a technical background to comprehend their internal mechanisms and use these powerful models more consciously and responsibly.
Our focus for future works is on providing better decoding and manipulation of hidden states, with particular attention on information injection. We are working to integrate causal analysis via mechanistic interpretabilty to better explain the ``causes'' of a certain output.

%%\section*{References}
%\begin{thebibliography}{00}
%\bibitem{b1} G. Eason, B. Noble, and I. N. Sneddon, ``On certain integrals of Lipschitz-Hankel type involving products of Bessel functions,'' Phil. Trans. Roy. Soc. London, vol. A247, pp. 529--551, April 1955.
%\bibitem{b2} J. Clerk Maxwell, A Treatise on Electricity and Magnetism, 3rd ed., vol. 2. Oxford: Clarendon, 1892, pp.68--73.
%\bibitem{b3} I. S. Jacobs and C. P. Bean, ``Fine particles, thin films and exchange anisotropy,'' in Magnetism, vol. III, G. T. Rado and H. Suhl, Eds. New York: Academic, 1963, pp. 271--350.
%\bibitem{b4} K. Elissa, ``Title of paper if known,'' unpublished.
%\bibitem{b5} R. Nicole, ``Title of paper with only first word capitalized,'' J. Name Stand. Abbrev., in press.
%\bibitem{b6} Y. Yorozu, M. Hirano, K. Oka, and Y. Tagawa, ``Electron spectroscopy studies on magneto-optical media and plastic substrate interface,'' IEEE Transl. J. Magn. Japan, vol. 2, pp. 740--741, August 1987 [Digests 9th Annual Conf. Magnetics Japan, p. 301, 1982].
%\bibitem{b7} M. Young, The Technical Writer's Handbook. Mill Valley, CA: University Science, 1989.
%\end{thebibliography}
%\vspace{12pt}
%\color{red}
%IEEE conference templates contain guidance text for composing and formatting conference papers. Please ensure that all template text is removed from your conference paper prior to submission to the conference. Failure to remove the template text from your paper may result in your paper not being published.

\bibliographystyle{IEEEtran}
\bibliography{bibliography}

\begin{thebibliography}{10}
\providecommand{\url}[1]{#1}
\csname url@samestyle\endcsname
\providecommand{\newblock}{\relax}
\providecommand{\bibinfo}[2]{#2}
\providecommand{\BIBentrySTDinterwordspacing}{\spaceskip=0pt\relax}
\providecommand{\BIBentryALTinterwordstretchfactor}{4}
\providecommand{\BIBentryALTinterwordspacing}{\spaceskip=\fontdimen2\font plus
\BIBentryALTinterwordstretchfactor\fontdimen3\font minus \fontdimen4\font\relax}
\providecommand{\BIBforeignlanguage}[2]{{%
\expandafter\ifx\csname l@#1\endcsname\relax
\typeout{** WARNING: IEEEtran.bst: No hyphenation pattern has been}%
\typeout{** loaded for the language `#1'. Using the pattern for}%
\typeout{** the default language instead.}%
\else
\language=\csname l@#1\endcsname
\fi
#2}}
\providecommand{\BIBdecl}{\relax}
\BIBdecl

\bibitem{DBLP:journals/npjdm/MehandruMASBA24}
\BIBentryALTinterwordspacing
N.~Mehandru, B.~Y. Miao, E.~R. Almaraz, M.~Sushil, A.~J. Butte, and A.~M. Alaa, ``Evaluating large language models as agents in the clinic,'' \emph{npj Digit. Medicine}, vol.~7, no.~1, 2024. [Online]. Available: \url{https://doi.org/10.1038/s41746-024-01083-y}
\BIBentrySTDinterwordspacing

\bibitem{DBLP:journals/corr/abs-2303-17564}
\BIBentryALTinterwordspacing
S.~Wu, O.~Irsoy, S.~Lu, V.~Dabravolski, M.~Dredze, S.~Gehrmann \emph{et~al.}, ``Bloomberggpt: {A} large language model for finance,'' \emph{CoRR}, vol. abs/2303.17564, 2023. [Online]. Available: \url{https://doi.org/10.48550/arXiv.2303.17564}
\BIBentrySTDinterwordspacing

\bibitem{DBLP:journals/corr/abs-2203-15556}
\BIBentryALTinterwordspacing
J.~Hoffmann, S.~Borgeaud, A.~Mensch, E.~Buchatskaya, T.~Cai, E.~Rutherford \emph{et~al.}, ``Training compute-optimal large language models,'' \emph{CoRR}, vol. abs/2203.15556, 2022. [Online]. Available: \url{https://doi.org/10.48550/arXiv.2203.15556}
\BIBentrySTDinterwordspacing

\bibitem{DBLP:journals/corr/abs-2303-08774}
\BIBentryALTinterwordspacing
OpenAI, ``{GPT-4} technical report,'' \emph{CoRR}, vol. abs/2303.08774, 2023. [Online]. Available: \url{https://doi.org/10.48550/arXiv.2303.08774}
\BIBentrySTDinterwordspacing

\bibitem{DBLP:conf/nips/VaswaniSPUJGKP17}
\BIBentryALTinterwordspacing
A.~Vaswani, N.~Shazeer, N.~Parmar, J.~Uszkoreit, L.~Jones, A.~N. Gomez \emph{et~al.}, ``Attention is all you need,'' in \emph{Advances in Neural Information Processing Systems 30: Annual Conference on Neural Information Processing Systems 2017, December 4-9, 2017, Long Beach, CA, {USA}}, I.~Guyon, U.~von Luxburg, S.~Bengio, H.~M. Wallach, R.~Fergus, S.~V.~N. Vishwanathan, and R.~Garnett, Eds., 2017, pp. 5998--6008. [Online]. Available: \url{https://proceedings.neurips.cc/paper/2017/hash/3f5ee243547dee91fbd053c1c4a845aa-Abstract.html}
\BIBentrySTDinterwordspacing

\bibitem{DBLP:journals/csur/JiLFYSXIBMF23}
\BIBentryALTinterwordspacing
Z.~Ji, N.~Lee, R.~Frieske, T.~Yu, D.~Su, Y.~Xu \emph{et~al.}, ``Survey of hallucination in natural language generation,'' \emph{{ACM} Comput. Surv.}, vol.~55, no.~12, pp. 248:1--248:38, 2023. [Online]. Available: \url{https://doi.org/10.1145/3571730}
\BIBentrySTDinterwordspacing

\bibitem{DBLP:conf/nips/LiuAGKZ23}
\BIBentryALTinterwordspacing
B.~Liu, J.~T. Ash, S.~Goel, A.~Krishnamurthy, and C.~Zhang, ``Exposing attention glitches with flip-flop language modeling,'' in \emph{Advances in Neural Information Processing Systems 36: Annual Conference on Neural Information Processing Systems 2023, NeurIPS 2023, New Orleans, LA, USA, December 10 - 16, 2023}, A.~Oh, T.~Naumann, A.~Globerson, K.~Saenko, M.~Hardt, and S.~Levine, Eds., 2023. [Online]. Available: \url{http://papers.nips.cc/paper\_files/paper/2023/hash/510ad3018bbdc5b6e3b10646e2e35771-Abstract-Conference.html}
\BIBentrySTDinterwordspacing

\bibitem{DBLP:journals/corr/abs-2207-09238}
\BIBentryALTinterwordspacing
M.~Phuong and M.~Hutter, ``Formal algorithms for transformers,'' \emph{CoRR}, vol. abs/2207.09238, 2022. [Online]. Available: \url{https://doi.org/10.48550/arXiv.2207.09238}
\BIBentrySTDinterwordspacing

\bibitem{radford2019language}
A.~Radford, J.~Wu, R.~Child, D.~Luan, D.~Amodei, and I.~Sutskever, ``Language models are unsupervised multitask learners,'' \emph{{OpenAI Blog}}, 2019.

\bibitem{DBLP:journals/corr/abs-2403-08295}
\BIBentryALTinterwordspacing
T.~Mesnard, C.~Hardin, R.~Dadashi, S.~Bhupatiraju, S.~Pathak, L.~Sifre \emph{et~al.}, ``Gemma: Open models based on gemini research and technology,'' \emph{CoRR}, vol. abs/2403.08295, 2024. [Online]. Available: \url{https://doi.org/10.48550/arXiv.2403.08295}
\BIBentrySTDinterwordspacing

\bibitem{DBLP:journals/corr/abs-2310-06825}
\BIBentryALTinterwordspacing
A.~Q. Jiang, A.~Sablayrolles, A.~Mensch, C.~Bamford, D.~S. Chaplot, D.~de~Las~Casas \emph{et~al.}, ``Mistral 7b,'' \emph{CoRR}, vol. abs/2310.06825, 2023. [Online]. Available: \url{https://doi.org/10.48550/arXiv.2310.06825}
\BIBentrySTDinterwordspacing

\bibitem{DBLP:journals/corr/abs-2307-09288}
\BIBentryALTinterwordspacing
H.~Touvron, L.~Martin, K.~Stone, P.~Albert, A.~Almahairi, Y.~Babaei \emph{et~al.}, ``Llama 2: Open foundation and fine-tuned chat models,'' \emph{CoRR}, vol. abs/2307.09288, 2023. [Online]. Available: \url{https://doi.org/10.48550/arXiv.2307.09288}
\BIBentrySTDinterwordspacing

\bibitem{DBLP:journals/corr/abs-2303-08112}
\BIBentryALTinterwordspacing
N.~Belrose, Z.~Furman, L.~Smith, D.~Halawi, I.~Ostrovsky, L.~McKinney \emph{et~al.}, ``Eliciting latent predictions from transformers with the tuned lens,'' \emph{CoRR}, vol. abs/2303.08112, 2023. [Online]. Available: \url{https://doi.org/10.48550/arXiv.2303.08112}
\BIBentrySTDinterwordspacing

\bibitem{DBLP:journals/corr/abs-2405-00208}
\BIBentryALTinterwordspacing
J.~Ferrando, G.~Sarti, A.~Bisazza, and M.~R. Costa{-}juss{\`{a}}, ``A primer on the inner workings of transformer-based language models,'' \emph{CoRR}, vol. abs/2405.00208, 2024. [Online]. Available: \url{https://doi.org/10.48550/arXiv.2405.00208}
\BIBentrySTDinterwordspacing

\bibitem{nostalgebraist2020logitlens}
\BIBentryALTinterwordspacing
nostalgebraist. (2020) Interpreting gpt: the logit lens. [Online]. Available: \url{https://www.lesswrong.com/posts/AcKRB8wDpdaN6v6ru/interpreting-gpt-the-logit-lens}
\BIBentrySTDinterwordspacing

\bibitem{DBLP:journals/corr/abs-2310-16270}
\BIBentryALTinterwordspacing
M.~Sakarvadia, A.~Khan, A.~Ajith, D.~Grzenda, N.~Hudson, A.~Bauer \emph{et~al.}, ``Attention lens: {A} tool for mechanistically interpreting the attention head information retrieval mechanism,'' \emph{CoRR}, vol. abs/2310.16270, 2023. [Online]. Available: \url{https://doi.org/10.48550/arXiv.2310.16270}
\BIBentrySTDinterwordspacing

\bibitem{DBLP:conf/conll/PalSYWB23}
\BIBentryALTinterwordspacing
K.~Pal, J.~Sun, A.~Yuan, B.~C. Wallace, and D.~Bau, ``Future lens: Anticipating subsequent tokens from a single hidden state,'' in \emph{Proceedings of the 27th Conference on Computational Natural Language Learning, CoNLL 2023, Singapore, December 6-7, 2023}, J.~Jiang, D.~Reitter, and S.~Deng, Eds.\hskip 1em plus 0.5em minus 0.4em\relax Association for Computational Linguistics, 2023, pp. 548--560. [Online]. Available: \url{https://doi.org/10.18653/v1/2023.conll-1.37}
\BIBentrySTDinterwordspacing

\bibitem{DBLP:conf/iclr/AlainB17}
\BIBentryALTinterwordspacing
G.~Alain and Y.~Bengio, ``Understanding intermediate layers using linear classifier probes,'' in \emph{5th International Conference on Learning Representations, {ICLR} 2017, Toulon, France, April 24-26, 2017, Workshop Track Proceedings}.\hskip 1em plus 0.5em minus 0.4em\relax OpenReview.net, 2017. [Online]. Available: \url{https://openreview.net/forum?id=HJ4-rAVtl}
\BIBentrySTDinterwordspacing

\bibitem{DBLP:conf/acl/Vig19}
\BIBentryALTinterwordspacing
J.~Vig, ``A multiscale visualization of attention in the transformer model,'' in \emph{Proceedings of the 57th Conference of the Association for Computational Linguistics, {ACL} 2019, Florence, Italy, July 28 - August 2, 2019, Volume 3: System Demonstrations}, M.~R. Costa{-}juss{\`{a}} and E.~Alfonseca, Eds.\hskip 1em plus 0.5em minus 0.4em\relax Association for Computational Linguistics, 2019, pp. 37--42. [Online]. Available: \url{https://doi.org/10.18653/v1/p19-3007}
\BIBentrySTDinterwordspacing

\bibitem{DBLP:conf/acl/WangTC21}
\BIBentryALTinterwordspacing
Z.~J. Wang, R.~Turko, and D.~H. Chau, ``Dodrio: Exploring transformer models with interactive visualization,'' in \emph{Proceedings of the Joint Conference of the 59th Annual Meeting of the Association for Computational Linguistics and the 11th International Joint Conference on Natural Language Processing, {ACL} 2021 - System Demonstrations, Online, August 1-6, 2021}, H.~Ji, J.~C. Park, and R.~Xia, Eds.\hskip 1em plus 0.5em minus 0.4em\relax Association for Computational Linguistics, 2021, pp. 132--141. [Online]. Available: \url{https://doi.org/10.18653/v1/2021.acl-demo.16}
\BIBentrySTDinterwordspacing

\bibitem{DBLP:journals/corr/abs-2404-07004}
\BIBentryALTinterwordspacing
I.~Tufanov, K.~Hambardzumyan, J.~Ferrando, and E.~Voita, ``{LM} transparency tool: Interactive tool for analyzing transformer language models,'' \emph{CoRR}, vol. abs/2404.07004, 2024. [Online]. Available: \url{https://doi.org/10.48550/arXiv.2404.07004}
\BIBentrySTDinterwordspacing

\bibitem{elhage2021mathematical}
\BIBentryALTinterwordspacing
N.~Elhage, N.~Nanda, C.~Olsson, T.~Henighan, N.~Joseph, B.~Mann \emph{et~al.} (2021) A mathematical framework for transformer circuits. [Online]. Available: \url{https://transformer-circuits.pub/2021/framework/index.html}
\BIBentrySTDinterwordspacing

\bibitem{DBLP:conf/iclr/WangVCSS23}
\BIBentryALTinterwordspacing
K.~R. Wang, A.~Variengien, A.~Conmy, B.~Shlegeris, and J.~Steinhardt, ``Interpretability in the wild: a circuit for indirect object identification in {GPT-2} small,'' in \emph{The Eleventh International Conference on Learning Representations, {ICLR} 2023, Kigali, Rwanda, May 1-5, 2023}.\hskip 1em plus 0.5em minus 0.4em\relax OpenReview.net, 2023. [Online]. Available: \url{https://openreview.net/forum?id=NpsVSN6o4ul}
\BIBentrySTDinterwordspacing

\bibitem{DBLP:conf/kdd/Ribeiro0G16}
\BIBentryALTinterwordspacing
M.~T. Ribeiro, S.~Singh, and C.~Guestrin, ``"why should {I} trust you?": Explaining the predictions of any classifier,'' in \emph{Proceedings of the 22nd {ACM} {SIGKDD} International Conference on Knowledge Discovery and Data Mining, San Francisco, CA, USA, August 13-17, 2016}, B.~Krishnapuram, M.~Shah, A.~J. Smola, C.~C. Aggarwal, D.~Shen, and R.~Rastogi, Eds.\hskip 1em plus 0.5em minus 0.4em\relax {ACM}, 2016, pp. 1135--1144. [Online]. Available: \url{https://doi.org/10.1145/2939672.2939778}
\BIBentrySTDinterwordspacing

\bibitem{DBLP:conf/nips/GeigerLIP21}
\BIBentryALTinterwordspacing
A.~Geiger, H.~Lu, T.~Icard, and C.~Potts, ``Causal abstractions of neural networks,'' in \emph{Advances in Neural Information Processing Systems 34: Annual Conference on Neural Information Processing Systems 2021, NeurIPS 2021, December 6-14, 2021, virtual}, M.~Ranzato, A.~Beygelzimer, Y.~N. Dauphin, P.~Liang, and J.~W. Vaughan, Eds., 2021, pp. 9574--9586. [Online]. Available: \url{https://proceedings.neurips.cc/paper/2021/hash/4f5c422f4d49a5a807eda27434231040-Abstract.html}
\BIBentrySTDinterwordspacing

\bibitem{DBLP:journals/corr/abs-2401-06102}
\BIBentryALTinterwordspacing
A.~Ghandeharioun, A.~Caciularu, A.~Pearce, L.~Dixon, and M.~Geva, ``Patchscopes: {A} unifying framework for inspecting hidden representations of language models,'' \emph{CoRR}, vol. abs/2401.06102, 2024. [Online]. Available: \url{https://doi.org/10.48550/arXiv.2401.06102}
\BIBentrySTDinterwordspacing

\bibitem{elhage2021garcon}
\BIBentryALTinterwordspacing
N.~Elhage, N.~Nanda, C.~Olsson, T.~Henighan, N.~Joseph, B.~Mann \emph{et~al.} (2021) Transformerlens. [Online]. Available: \url{https://transformer-circuits.pub/2021/garcon/index.html}
\BIBentrySTDinterwordspacing

\bibitem{nanda2022transformerlens}
N.~Nanda and J.~Bloom, ``Transformerlens,'' \url{https://github.com/TransformerLensOrg/TransformerLens}, 2022.

\bibitem{DBLP:conf/nips/0001LV23}
\BIBentryALTinterwordspacing
M.~Hanna, O.~Liu, and A.~Variengien, ``How does {GPT-2} compute greater-than?: Interpreting mathematical abilities in a pre-trained language model,'' in \emph{Advances in Neural Information Processing Systems 36: Annual Conference on Neural Information Processing Systems 2023, NeurIPS 2023, New Orleans, LA, USA, December 10 - 16, 2023}, A.~Oh, T.~Naumann, A.~Globerson, K.~Saenko, M.~Hardt, and S.~Levine, Eds., 2023. [Online]. Available: \url{http://papers.nips.cc/paper\_files/paper/2023/hash/efbba7719cc5172d175240f24be11280-Abstract-Conference.html}
\BIBentrySTDinterwordspacing

\bibitem{DBLP:conf/nips/ConmyMLHG23}
\BIBentryALTinterwordspacing
A.~Conmy, A.~N. Mavor{-}Parker, A.~Lynch, S.~Heimersheim, and A.~Garriga{-}Alonso, ``Towards automated circuit discovery for mechanistic interpretability,'' in \emph{Advances in Neural Information Processing Systems 36: Annual Conference on Neural Information Processing Systems 2023, NeurIPS 2023, New Orleans, LA, USA, December 10 - 16, 2023}, A.~Oh, T.~Naumann, A.~Globerson, K.~Saenko, M.~Hardt, and S.~Levine, Eds., 2023. [Online]. Available: \url{http://papers.nips.cc/paper\_files/paper/2023/hash/34e1dbe95d34d7ebaf99b9bcaeb5b2be-Abstract-Conference.html}
\BIBentrySTDinterwordspacing

\end{thebibliography}

\end{document}